\documentclass{article}
\usepackage{spconf,amsmath,graphicx}
\usepackage{cleveref}
\usepackage{booktabs}  
\usepackage{graphicx}  
\usepackage{multirow}  
\usepackage{enumitem}
\usepackage{geometry}
\setlist{nosep, leftmargin=14pt}
\usepackage{adjustbox} 
\usepackage{mwe} 
\usepackage{amssymb}
\usepackage{url} 

\title{XBench: A Comprehensive Benchmark for Visual-Language \\Explanations in Chest Radiography}
\name{Haozhe Luo$^{1,3}$ , Shelley Zixin Shu$^{1}$, Ziyu Zhou$^{2}$, Sebastian Otálora$^{3}$, Mauricio Reyes$^{1,4}$}
\address{ $^{1}$ARTORG Center for Biomedical Engineering Research, University of Bern, Switzerland\\$^{2}$Shanghai Jiao Tong University, China \\ $^{3}$Kaiko.AI, Switzerland \\  $^{4}$Dept. of Radiation Oncology, Inselspital, Bern University Hospital and \\ University of Bern, Switzerland}

\begin{document}
%
\maketitle

\section{Abstract}
Vision–language models (VLMs) have recently shown remarkable zero-shot performance in medical image understanding, yet their grounding ability, the extent to which textual concepts align with visual evidence, remains underexplored. In the medical domain, however, reliable grounding is essential for interpretability and clinical adoption. In this work, we present the first systematic benchmark for evaluating cross-modal interpretability in chest X-rays across seven CLIP-style VLM variants. We generate visual explanations using cross-attention and similarity-based localization maps, and quantitatively assess their alignment with radiologist-annotated regions across multiple pathologies. Our analysis reveals that: (1) while all VLM variants demonstrate reasonable localization for large and well-defined pathologies, their performance substantially degrades for small or diffuse lesions; (2) models that are pretrained on chest X-ray–specific datasets exhibit improved alignment compared to those trained on general-domain data. (3) The overall recognition ability and grounding ability of the model are strongly correlated.
These findings underscore that current VLMs, despite their strong recognition ability, still fall short in clinically reliable grounding, highlighting the need for targeted interpretability benchmarks before deployment in medical practice. \textsc{XBench} code is available at https://github.com/Roypic/Benchmarkingattention.

\begin{keywords}
Explainability, Benchmark, VLM, Grounding
\end{keywords}

\begin{figure*}[t!]
    \centering
    \includegraphics[width=16cm]{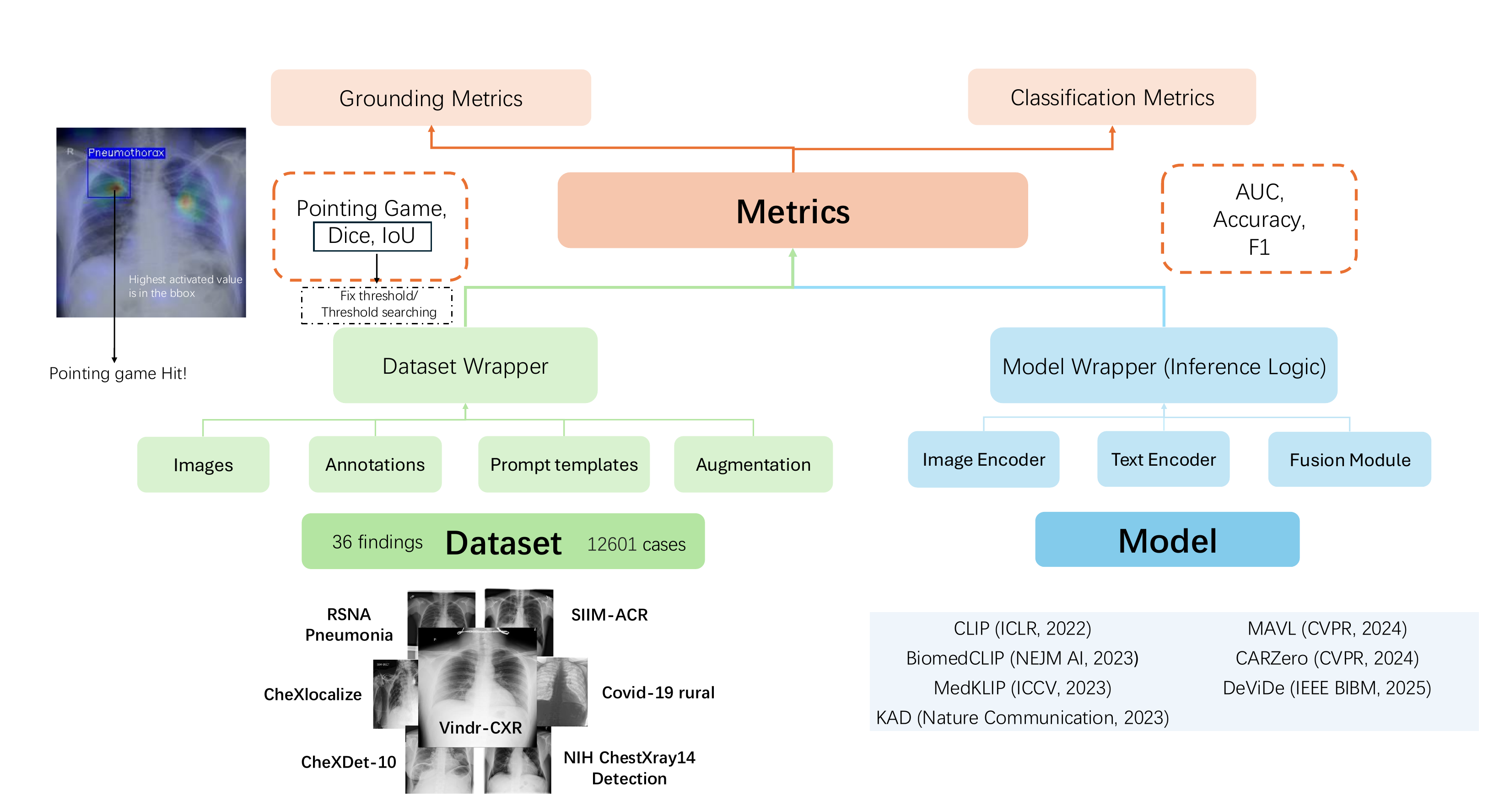}
    \caption{Overview of the unified evaluation framework for medical vision-language models. 
    The framework integrates three main components: \textbf{Dataset}, \textbf{Model}, and \textbf{Metrics}. 
    The \textbf{Dataset Wrapper} organizes multi-source datasets (36 diseases, 12,601 cases) including images, annotations, prompt templates, and augmentations from RSNA Pneumonia\cite{rsna}, Covid19-rural\cite{desai2020chest}, ChestDet-10\cite{liu2020chestx}, SIIM\cite{siim-acr-pneumothorax-segmentation}, CheXlocalize\cite{saporta2022benchmarking}, ChestXray14 Detection\cite{wang2017chestx}, and Vindr-CXR\cite{nguyen2022vindr}. 
    The \textbf{Model Wrapper} standardizes inference logic across vision-language models, encapsulating the image encoder, text encoder, and fusion module; it supports representative models such as CLIP, BiomedCLIP, MedKILP, KAD, MAVL, CARZero, and DeViDe. 
    The \textbf{Metrics} module unifies both \textbf{Grounding Metrics} (e.g., Pointing Game, Dice, IoU with fixed or searched thresholds) and \textbf{Classification Metrics} (e.g., AUC, Accuracy, F1), enabling comprehensive and comparable evaluation across datasets and models.}
    
    \label{fig:framework}
\end{figure*}
\section{Introduction}
Deep learning has achieved remarkable progress in medical image analysis, enabling automated interpretation of chest radiographs at expert-level accuracy in certain diagnostic tasks. Recent advances in vision-language models (VLMs) \cite{zhang2023knowledge,luo2024devide, wang2022medclip} further extend this capability by jointly learning from paired image-text data, demonstrating strong zero-shot recognition and transferability across medical domains. However, in clinical settings, the value of such models extends beyond classification accuracy. Whether models’ predictions are grounded in meaningful visual evidence are equally important \cite{luo2024dwarf, pahud2024orchestrating, saporta2022benchmarking}, as reliable grounding, or cross-modal interpretability, is essential for clinical trust, model validation, and regulatory acceptance.

While numerous VLMs\cite{radford2021learning,zhang2023biomedclip,wu2023medklip} have shown promising performance on image-level tasks, their spatial reasoning and localization abilities remain poorly understood. Prior work has revealed that post-hoc saliency methods, though widely adopted for medical explainability, often fail to localize fine-grained or small-scale pathologies compared with radiologist annotations~\cite{saporta2022benchmarking}. In particular, benchmarks like CheXlocalize\cite{saporta2022benchmarking} have highlighted large gaps between model-generated heatmaps and expert-drawn regions, underscoring the need for standardized and quantitative evaluation of grounding performance.

To address this gap, we introduce \textbf{XBench}, the first comprehensive benchmark for evaluating cross-modal interpretability in chest X-rays. \textsc{XBench} integrates the \textbf{Dataset}, \textbf{Model}, and \textbf{Metrics} modules into a unified evaluation framework (Fig.~\ref{fig:framework}), supporting seven representative CLIP-style VLMs spanning pretraining from natural images to chest X-ray–specific data. Grounding performance is assessed with Pointing Game, Dice, and IoU, while AUC, Accuracy, and F1 are jointly reported. Across 36 findings and 12{,}601 cases, \textsc{XBench} reveals systematic explainability patterns: domain-specific pretraining improves grounding for large, well-defined pathologies, but models still underperform on small lesions, obfuscated/overlapping regions, and diffuse or scale-variant findings.

Together, this benchmark establishes a rigorous foundation for evaluating and improving the interpretability of medical vision--language models, paving the way toward clinically reliable multimodal AI.

\begin{table*}[t]
\centering
\setlength{\tabcolsep}{3pt}
\resizebox{\linewidth}{!}{
\begin{tabular}{l|ccc|ccc|ccc|ccc|ccc|ccc|ccc}
\toprule
\textbf{Method} &
\multicolumn{3}{c|}{\textbf{COVID-19}} &
\multicolumn{3}{c|}{\textbf{Pneumonia}} &
\multicolumn{3}{c|}{\textbf{Pneumothorax}} &
\multicolumn{3}{c|}{\textbf{CheXDet-10}} &
\multicolumn{3}{c|}{\textbf{CheXlocalize}} &
\multicolumn{3}{c|}{\textbf{VinDR-CXR}} &
\multicolumn{3}{c}{\textbf{ChestX-ray14}} \\
\midrule
 & Point & Dice & IoU & Point & Dice & IoU & Point & Dice & IoU & Point & Dice & IoU & Point & Dice & IoU & Point & Dice & IoU & Point & Dice & IoU \\
\midrule
CLIP\cite{radford2021learning}        & 15.62 & 18.31 & 10.92 & 7.39 & 20.20 & 11.97 & 0.46 & 3.16 & 1.64 & 7.10 & 12.99 & 8.10 & 3.22 & 7.96 & 4.33 & 2.47 & 7.31 & 4.25 & 7.42 & 12.36 & 7.11 \\
BiomedCLIP\cite{zhang2023biomedclip}  & 3.12 & 16.56 & 9.69 & 12.64 & 20.20 & 11.97 & 1.01 & 3.17 & 1.64 & 5.41 & 12.68 & 7.92 & 3.24 & 8.16 & 4.46 & 3.95 & 7.55 & 4.39 & 9.05 & 12.67 & 7.37 \\
MedKLIP\cite{wu2023medklip}     & \underline{28.12} & 19.25 & 11.24 & 42.78 & 33.07 & 21.25 & 1.55 & 3.79 & 2.00 & 31.79 & 23.32 & 15.38 & 18.12 & 17.91 & 10.98 & \underline{23.74} & 19.04 & 12.02 & 40.06 & 27.74 & 18.21 \\
KAD\cite{zhang2023knowledge}         & 6.25 & 27.45 & 18.18 & 70.11 & \underline{42.06} & \underline{28.05} & 2.47 & 4.18 & 2.20 & 32.18 & 23.26 & 15.47 & 24.21 & \underline{18.73} & \underline{11.61} & 18.35 & 18.55 & 11.80 & 43.61 & 29.65 & 19.41 \\
MAVL\cite{phan2024decomposing}        & 15.62 & 16.45 & 9.61 & 29.31 & 20.14 & 11.94 & 2.78 & 4.09 & 2.25 & 26.12 & 19.19 & 12.58 & 17.49 & 13.31 & 7.88 & 15.89 & 16.53 & 10.32 & 31.31 & 20.83 & 13.12 \\
CARZero\cite{lai2024carzero}     & \textbf{53.12} & \textbf{36.64} & \textbf{24.26} & \textbf{83.66} & \textbf{50.47} & \textbf{36.45} & \textbf{5.56} & \textbf{4.94} & \textbf{2.79} & \textbf{48.38} & \textbf{31.35} & \textbf{22.40} & \textbf{33.35} & \textbf{23.20} & \textbf{15.54} & \textbf{39.07} & \textbf{31.01} & \textbf{22.28} & \textbf{61.57} & \textbf{39.44} & \textbf{28.01} \\
DeViDe\cite{luo2024devide}      & 3.12 & \underline{28.36} & \underline{18.56} & \underline{70.77} & 40.16 & 26.48 & \underline{3.40} & \underline{4.36} & \underline{2.33} & \underline{35.26} & \underline{26.56} & \underline{18.12} & \underline{27.02} & 18.22 & 11.22 & 21.47 & \underline{20.43} & \underline{13.03} & \underline{49.16} & \underline{30.57} & \underline{20.04} \\
\bottomrule
\end{tabular}
}
\caption{\textbf{Grounding metrics across datasets.} Each dataset group shows the mean Pointing Game and the best-threshold Dice/IoU. Single-disease datasets focus on one pathology; multi-disease show averages over classes. All values are percentages. Best and second-best in each column are in \textbf{bold} and \underline{underline}, respectively.}
\label{tab:grounding_all_diseases}
\end{table*}

\begin{table*}[t]
\centering
\caption{\textbf{Per-class Pointing Game performance on VinDR-CXR.} Each model spans two rows to display all classes. The best result per class is shown in \textbf{bold}, and the second best is \underline{underlined}. For most findings, the grounding performance of all models remains below 50\%.}
\resizebox{0.95\textwidth}{!}{%
\begin{tabular}{lcccccccccccc}
\toprule
\multirow{4}{*}{\textbf{Model}} & \multirow{4}{*}{\textbf{Mean}} & \multicolumn{11}{c}{\textbf{Classes}} \\
\cmidrule(lr){3-13}
 &  & \textbf{Aortic enl.} & \textbf{Atelectasis} & \textbf{Calcif.} & \textbf{Cardiomegaly} & \textbf{Clav. fract.} & \textbf{Consol.} & \textbf{Emphysema} & \textbf{Enl. PA} & \textbf{ILD} & \textbf{Infiltration} & \textbf{Lung Opac.} \\
\cmidrule(lr){3-13}
 &  & \textbf{Lung cavity} & \textbf{Lung cyst} & \textbf{Mediast. shift} & \textbf{Nodule/Mass} & \textbf{Other lesion} & \textbf{Pleural eff.} & \textbf{Pleural thick.} & \textbf{Pneumothorax} & \textbf{Pulm. fibrosis} & \textbf{Rib fract.} &  \\
\midrule
\multirow{2}{*}{CLIP} & \multirow{2}{*}{2.47} & 5.73 & 0.00 & 0.60 & 9.26 & 0.00 & 6.90 & 0.00 & 0.00 & 15.10 & 1.89 & 0.00 \\
 &  & 0.00 & \underline{0.00} & 0.00 & 1.34 & 6.17 & 4.26 & 0.66 & 0.00 & 0.00 & 0.00 \\[2pt]
\midrule
\multirow{2}{*}{BioMedCLIP} & \multirow{2}{*}{3.95} & 7.73 & 1.20 & 0.00 & 7.90 & 0.00 & 0.00 & 0.00 & 0.00 & 2.80 & 3.51 & 0.00 \\
 &  & 0.00 & \textbf{50.00} & 0.00 & 1.25 & 1.12 & 4.85 & 0.62 & 0.00 & 1.91 & 0.00 \\[2pt]
\midrule
\multirow{2}{*}{MedKLIP} & \multirow{2}{*}{23.74} & 30.21 & 24.68 & 7.14 & 58.52 & 0.00 & 50.57 & \textbf{33.33} & \textbf{42.86} & \textbf{40.62} & \underline{32.08} & \underline{32.00} \\
 &  & \textbf{25.00} & 0.00 & \underline{25.00} & 24.16 & \underline{18.52} & 14.89 & 0.00 & 20.00 & \underline{18.88} & 0.00 \\[2pt]
\midrule
\multirow{2}{*}{KAD} & \multirow{2}{*}{18.35} & 12.08 & 21.69 & 14.44 & \underline{86.60} & 0.00 & 62.37 & 0.00 & 0.00 & 1.40 & 19.30 & 27.50 \\
 &  & 0.00 & 0.00 & 0.00 & 34.38 & 3.37 & \underline{38.83} & \underline{2.47} & \underline{22.22} & 2.39 & \textbf{36.36} \\[2pt]
\midrule
\multirow{2}{*}{MAVL} & \multirow{2}{*}{15.89} & \underline{30.73} & 12.99 & 3.57 & 12.59 & 0.00 & 34.48 & \textbf{33.33} & 14.29 & \underline{33.85} & 18.87 & 26.67 \\
 &  & 12.50 & 0.00 & 6.25 & 9.40 & 18.52 & 5.32 & 0.00 & \textbf{33.33} & 17.86 & 9.09 \\[2pt]
\midrule
\multirow{2}{*}{CARZero} & \multirow{2}{*}{\textbf{39.07}} & \textbf{77.60} & \textbf{41.56} & \underline{26.19} & 76.30 & \textbf{100.00} & \textbf{77.01} & \textbf{33.33} & \underline{28.57} & 2.08 & \textbf{52.83} & \textbf{38.67} \\
 &  & \textbf{25.00} & 0.00 & \textbf{37.50} & \textbf{37.58} & \textbf{51.06} & 15.23 & \textbf{33.33} & \textbf{33.33} & \textbf{20.92} & \underline{27.27} \\[2pt]
\midrule
\multirow{2}{*}{DeViDe} & \multirow{2}{*}{21.47} & 10.63 & \underline{31.33} & \textbf{34.44} & \textbf{88.32} & 0.00 & \underline{73.12} & 0.00 & 14.29 & 2.80 & 17.54 & 31.25 \\
 &  & 0.00 & 0.00 & \textbf{37.50} & \underline{37.50} & 7.87 & \textbf{41.75} & 1.23 & 16.67 & 14.83 & \underline{27.27} \\[2pt]
\bottomrule
\end{tabular}%
}
\label{tab:vindr}
\end{table*}

\begin{table}[t]
\centering
\caption{Per-class Pointing game results on ChestX-ray14 dataset.}
\begin{adjustbox}{width=\linewidth}
\begin{tabular}{l|ccccccccc}
\toprule
\textbf{Model} & \textbf{Mean} & \textbf{ATE} & \textbf{CARD} & \textbf{EFF} & \textbf{INF} & \textbf{MASS} & \textbf{NOD} & \textbf{PNEU} & \textbf{PTX} \\
\midrule
CLIP      &  7.42 &  3.33 & 20.55 & 11.11 & 11.38 &  4.71 &  2.53        &  1.67 &  4.08 \\
BioMedCLIP&  9.05 &  2.78 & 42.47 &  0.65 &  4.88 &  7.06 &  0.0         & 12.5  &  2.04 \\
MedKLIP   & 40.06 & 32.78 & 81.51 & 26.14 & 51.22 & 35.29 & 11.39        & 56.67 & 25.51 \\
KAD       & 43.61 & 38.33 & \underline{90.41} & \underline{47.71} & 23.58 & 52.94 & 17.72        & \underline{60.83} & 17.35 \\
MAVL      & 31.31 & 33.33 & 52.74 &  9.8  & 40.65 & 37.65 &  7.59        & 35.0 & \underline{33.67} \\
CARZero   & \textbf{61.57} & \textbf{50.56} & \textbf{99.32} & \textbf{60.13} & \textbf{74.8} & \textbf{65.88} & \textbf{24.05} & \textbf{75.0} & \textbf{42.86} \\
DeViDe    & \underline{49.16} & \underline{43.33} & 91.78 & 44.44 & \underline{47.15} & \underline{54.12} & \textbf{24.05} & 70.0 & 18.37 \\

\bottomrule
\end{tabular}
\end{adjustbox}
\label{tab:chestxray14}
\end{table}

\begin{table}[t]
\centering
\caption{Per-class Pointing game results on CheXDet-10 dataset.}
\begin{adjustbox}{width=\linewidth}
\begin{tabular}{l|ccccccccccc}
\toprule
\textbf{Model} & \textbf{Mean} & \textbf{ATE} & \textbf{CALC} & \textbf{CONS} & \textbf{EFF} & \textbf{EMPH} & \textbf{FIB} & \textbf{FX} & \textbf{MASS} & \textbf{NOD} & \textbf{PTX} \\
\midrule
CLIP      & 7.11 & 27.08 & 0.00 & 6.14 & 7.00 & 10.81 & 18.67 & 0.00 & 0.00 & 1.35 & 0.00 \\
BioMedCLIP& 5.54 & 12.50 & 2.70 & 7.22 & 5.76 & 18.92 & 4.00 & 0.00 & 0.00 & 0.00 & 3.03 \\
MedKLIP   & 33.09 & 37.50 & \underline{5.41} & 67.15 & 40.74 & \textbf{59.46} & \underline{34.67} & 4.48 & 31.03 & 16.22 & \underline{21.21} \\
KAD       & 32.00 & 58.33 & 0.00 & \underline{74.73} & 59.26 & 32.43 & 8.00 & \underline{8.96} & 51.72 & 16.22 & 12.12 \\
MAVL      & 26.09 & 41.67 & 2.70 & 42.96 & 24.69 & \underline{45.95} & 30.67 & 0.00 & 51.72 & 2.70 & 18.18 \\
CARZero   & \textbf{48.96} & \textbf{66.67} & \textbf{14.29} & \textbf{81.01} & \textbf{70.74} & 61.76 & \textbf{45.71} & \textbf{20.31} & \textbf{64.29} & \textbf{23.29} & \textbf{35.71} \\
DeViDe    & \underline{34.27} & \textbf{66.67} & 0.00 & \textbf{78.34} & \underline{63.79} & 29.73 & 22.67 & 7.46 & \underline{58.62} & 16.22 & 9.09 \\

\bottomrule
\end{tabular}
\end{adjustbox}
\label{tab:chexdet10}
\end{table}

\begin{table}[t]
\centering
\caption{Per-class Pointing game results on CheXlocalize dataset.}
\begin{adjustbox}{width=\linewidth}
\begin{tabular}{lcccccccccccccc}
\toprule
\textbf{Model} & \textbf{Mean} & \textbf{Air. Opac.} & \textbf{ATE} & \textbf{CARD} & \textbf{CONS} & \textbf{EDEMA} & \textbf{Enl. CARD} & \textbf{FX} & \textbf{Lung Les.} & \textbf{EFF} & \textbf{Ple. Oth.} & \textbf{PNEU} & \textbf{PTX} & \textbf{Sup. Dev.} \\
\midrule
CLIP& 6.88 & 5.05 & 1.15 & 13.37 & 0.00 & 9.09 & 8.87 & 0.00 & 0.00 & 1.71 & 0.00 & 0.00 & 0.00 & 2.59 \\
BioMedCLIP & 4.49 & 1.44 & 1.72 & 13.95 & 0.00 & 3.90 & 8.53 & 0.00 & 0.00 & 0.00 & 0.00 & 10.00 & 0.00 & 2.59 \\
MedKLIP & 17.28 & \underline{33.57} & 4.60 & 34.88 & 3.45 & 22.08 & 39.93 & \textbf{16.67} & 21.43 & 5.98 & 0.00 & 40.00 & 0.00 & \underline{12.94} \\
KAD & 21.37 & 10.11 & 13.22 & \underline{73.84} & \underline{24.14} & 23.38 & \underline{51.54} & \textbf{16.67} & \textbf{42.86} & 15.38 & 0.00 & 40.00 & 0.00 & 3.56 \\
MAVL & 17.94 & 34.30 & 2.30 & 38.37 & 6.90 & 22.08 & 62.12 & \textbf{16.67} & 7.14 & 2.56 & 0.00 & 20.00 & 0.00 & 14.89 \\
CARZero & \textbf{33.38} & \textbf{40.43} & \textbf{14.37} & \textbf{86.63} & \textbf{37.93} & \textbf{35.06} & \textbf{66.55} & 0.00 & \textbf{42.86} & \textbf{23.08} & 0.00 & \textbf{50.00} & \textbf{18.18} & \textbf{18.45} \\
DeViDe & \underline{25.67} & 18.05 & \underline{16.67} & 77.91 & 27.59 & \underline{27.27} & 48.46 & \textbf{16.67} & 35.71 & \underline{15.38} & 0.00 & \textbf{50.00} & \underline{9.09} & 8.41 \\

\bottomrule
\end{tabular}
\end{adjustbox}
\label{tab:chexlocalize}
\end{table}


\section{Task Formulation \& Implementation Details}
We study zero-shot recognition and grounding of \(C\) diagnostic concepts \(\mathcal{C}\) over pooled datasets \(\mathcal{D}=\bigcup_{k=1}^{K}\mathcal{D}_k\). Each image \(x\in \mathbb{R}^{H\times W}\) has labels \(\mathbf{y}\in\{0,1\}^{C}\) and, when available, regions \(\mathcal{R}=\{R_c\subset\Omega\}_{c\in\mathcal{C}}\). A CLIP-style VLM \(f_\theta=(h_\theta,g_\theta)\) queries class  \(c\) via \(t_c=\tau(c)\) and computes \(s_c(x)=\langle h_\theta(x),g_\theta(t_c)\rangle\), \(p_c(x)=\sigma(s_c(x))\), and \(\hat y_c=\mathbb{I}\{p_c(x)\ge0.5\}\). Class saliency \(M_c(x)\) derives from similarity \(\phi_{\mathrm{sim}}\) or cross-attention \(\phi_{\mathrm{att}}\). Grounding uses normalized maps \(\widetilde{M}_c(x)\) thresholded as \(B_c(x;\tau)=\{u:\widetilde{M}_c(x)_u\ge\tau\}\). We compute Pointing Game \(\mathbb{I}\{\arg\max_u M_c(x)_u\in R_c^{(0.5)}\}\), Dice \(=\tfrac{2|B_c\cap R_c|}{|B_c|+|R_c|}\), and IoU \(=\tfrac{|B_c\cap R_c|}{|B_c\cup R_c|}\). Results are reported at fixed \((\gamma,\tau)=(0.5,0.5)\) and best \(\tau\), with recognition metrics (macro AUC, F1, AUPRC, Hamming acc) averaged per class and dataset. \\
We benchmark on seven CXR datasets: RSNA Pneumonia~\cite{rsna} (1 class), SIIM-ACR Pneumothorax~\cite{siim-acr-pneumothorax-segmentation} (1), COVID-19 Rural~\cite{desai2020chest} (1), CheXDet-10~\cite{liu2020chestx} (10), CheXlocalize~\cite{saporta2022benchmarking} (13), ChestX-ray14 Detection~\cite{wang2017chestx} (8), and VinDr-CXR~\cite{nguyen2022vindr} (21). Unless noted, models use batch size \(8\), input resolution \(224\times224\), and each model’s official prompt style for text encoding; grounding uses a best threshold searching strategy from 0 to 1 with step 0.01 (or a fixed threshold \(\tau=0.5\)). All experiments run on NVIDIA H200 GPUs (141\,GB). \textsc{XBench} supports custom component insertion, and only needs to modify the config file to achieve flexible reasoning.

\begin{figure*}[t!]
    \centering
    \includegraphics[width=16cm]{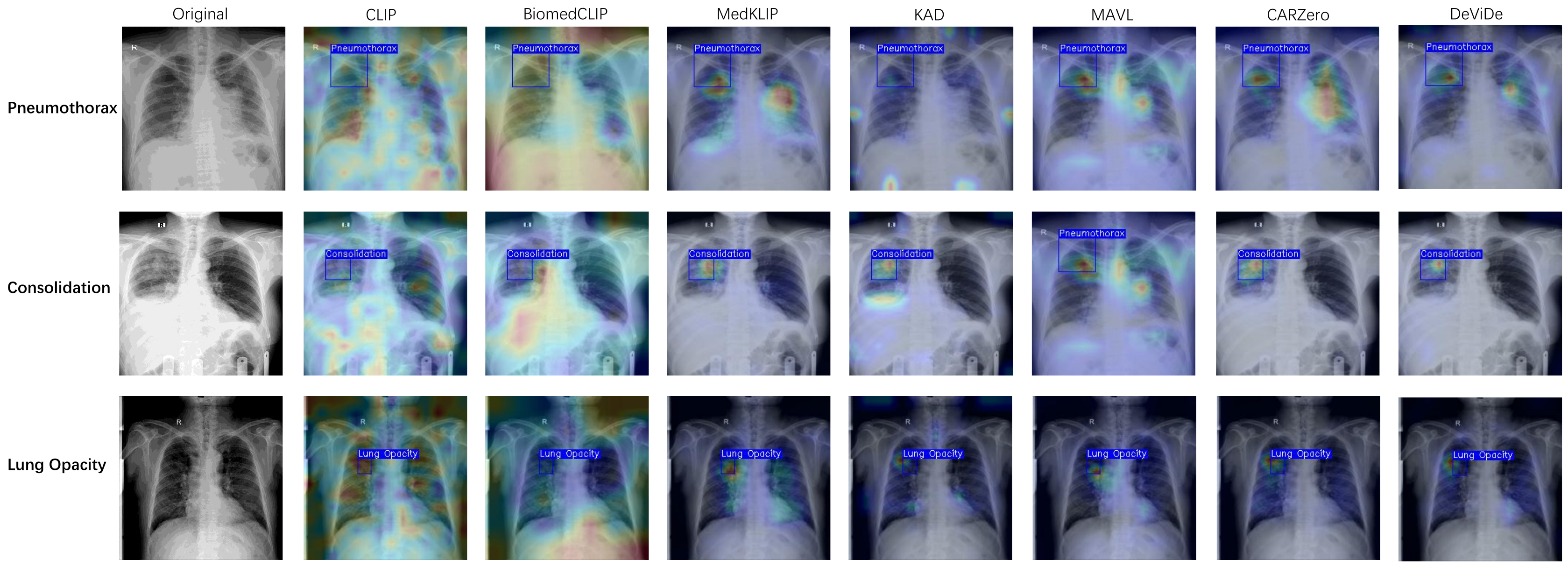}
    \caption{Visual comparison of disease localization across six vision-language models on chest X-rays. Blue boxes mark ground-truth regions. CARZero and DeViDe show more accurate and focused attention for Pneumothorax, Consolidation, and Lung Opacity.}
    \label{fig:attetionmap}
\end{figure*}

\begin{figure}[t!]
    \centering
    \includegraphics[width=8cm]{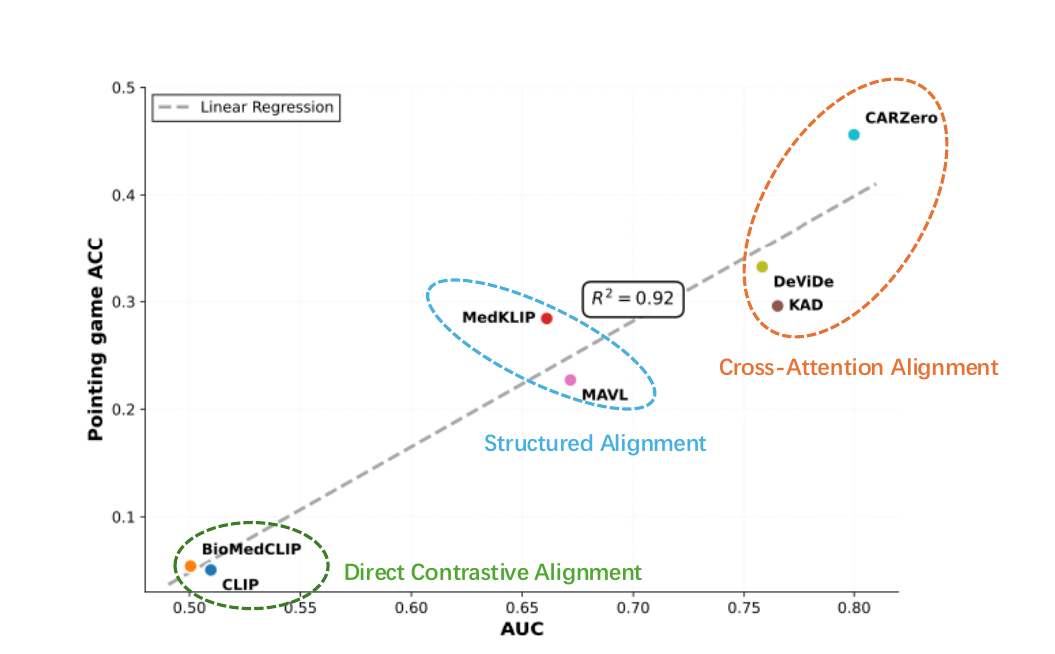}
    \caption{Correlation between disease classification and grounding accuracy across vision-language models.}
    \label{fig:pointinggamevsAUC}
\end{figure}

\begin{figure}[t!]
    \centering
    \includegraphics[width=8cm]{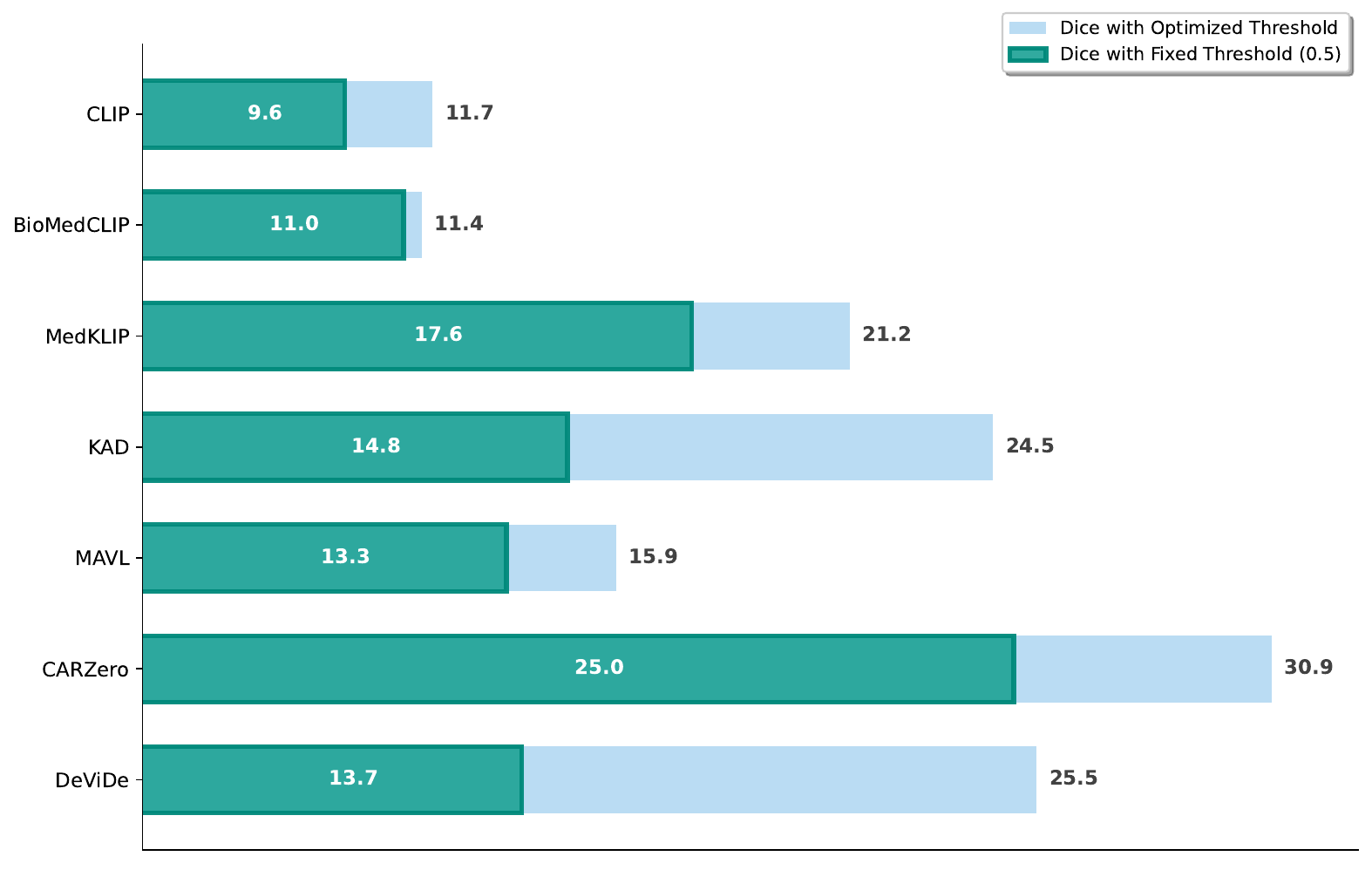}
    \caption{Comparison of Dice coefficients under fixed (0.5) and optimized thresholds across seven vision-language models. 
        CARZero achieves the highest Dice scores in both settings, indicating superior lesion localization consistency. }
    
    \label{fig:threshold}
\end{figure}

\section{Results and Analysis}
\subsubsection{Zero-Shot Classification Performance}
We evaluate seven CLIP-style VLMs in a zero-shot setting across multi- and single-disease datasets, emphasizing grounding without task-specific fine-tuning. As shown in Table.\ref{tab:grounding_all_diseases},\ref{tab:chestxray14},\ref{tab:chexdet10},\ref{tab:chexlocalize} and \ref{tab:vindr}, CARZero is consistently strongest. Gains are pronounced for large, well-defined findings (e.g., cardiomegaly, consolidation), all domain-sepcific models outperform natural-image baselines like CLIP, reflecting the value of CXR-specific pretraining. For emergent conditions such as COVID-19 (Table.\ref{tab:grounding_all_diseases}), CARZero also leads, substantially surpassing MedKLIP (0.19) and BioMedCLIP (0.30). Notably, for COVID-19 grounding and recognition, MAVL and MedKLIP outperform KAD and DeViDe, though their in-domain performance is lower—underscoring the importance of detailed query prompts at inference time. Corresponding Attention map visualization are shown in the Fig.\ref{fig:attetionmap}.

\subsubsection{Correlation Analysis and Transferability Insights}
As shown in Fig.~\ref{fig:pointinggamevsAUC}, \emph{classification} AUC and \emph{pointing-game} ACC are strongly coupled, as indicated by a high coefficient of determination ($R^2 = 0.92$): recognition gains typically strengthen grounding. Three pretraining regimes are: (i) direct contrastive alignment (CLIP, BioMedCLIP) with modest AUC and weak localization; (ii) structured alignment (MedKLIP, MAVL) in the mid-range; and (iii) cross-attention alignment (DeViDe, KAD, CARZero) in the upper-right with the best joint performance. Notably, CARZero lies above the trend, translating recognition performance into spatial evidence more efficiently. Overall, the monotonicity implies that strong classification performance often carries over to grounding.
\subsubsection{Threshold Sensitivity and Calibration Insights}
Fig.~\ref{fig:threshold} juxtaposes Dice scores at a fixed threshold ($\tau=0.5$) against threshold searched optimal value, highlighting calibration gaps $\Delta\text{Dice} = \text{Dice}_{\text{opt}} - \text{Dice}_{0.5}$. DeViDe and KAD show the largest discrepancies (11.8\%, 9.7\%), reflecting strong separability but skewed distributions near $\tau=0.5$; CARZero is moderate (5.9\%); MedKLIP, MAVL, and CLIP narrower (3.6\%, 2.6\%, 2.1\%); BioMedCLIP minimal (0.4\%). Smaller gaps enable deployment with little tuning, while larger ones demand post-hoc calibration or class-specific thresholds. Notably, high $\text{Dice}_\text{opt}$ models with big $\Delta\text{Dice}$ (e.g., DeViDe, KAD) pinpoint score calibration, not discriminability, as the key issue. Thus, report both fixed and optimized metrics, and prioritize calibration in tuning for better interpretability.
\subsubsection{Inconsistency between Grounding, Classification, and Recognition Difficulty}
A per-class analysis on VinDR-CXR uncovers that the intuitive correlation ``improved recognition yields enhanced grounding'' does not hold uniformly across pathologies. Notably, small or scale-variant lesions such as \textit{Pneumothorax}, \textit{Calcification}, and \textit{Nodule/Mass} reveal a stark recognition-grounding discrepancy: VLMs attain robust classification performance  but falter in providing faithful spatial cues (e.g., CARZero's Pointing Game performance are 0.33/0.26/0.38). Conversely, larger, shape-salient abnormalities like \textit{Cardiomegaly} elicit reliable localization (CARZero 0.76; DeViDe 0.88). Such patterns imply that prevailing VLMs excessively leverage global contextual priors while remaining vulnerable to lesion-scale ambiguities.

\section{Conclusion}
We present \textsc{XBench}, a unified benchmark for recognition and grounding in chest radiography. Across seven VLMs, we observe a strong model-level coupling between AUC and pointing accuracy, yet persistent per-class mismatches for small or scale-variant lesions, and notable calibration gaps between fixed and optimized thresholds. These results indicate that current medical VLMs still rely on global context and lack robust, size-aware spatial evidence. While our analysis centers on CLIP-style VLMs, recent domain-adapted MLLMs (e.g., a 7B LLaVA-Rad trained on ~697k radiograph–report pairs) have outperformed much larger general models (GPT-4V) on factual report generation. We'll further incorporaete such MLLM baselines in \textsc{XBench} to reveal whether their free-form explanations align better with radiologist annotations and how far foundation models have progressed.

\bibliographystyle{IEEEbib}
\bibliography{strings,refs}

\begin{thebibliography}{10}

\bibitem{rsna}
``Rsna pneumonia detection challenge (2018),'' \url{https://www.kaggle.com/c/rsna-pneumonia-detection-challenge}.

\bibitem{desai2020chest}
Shivang Desai, Ahmad Baghal, Thidathip Wongsurawat, Piroon Jenjaroenpun, Thomas Powell, Shaymaa Al-Shukri, Kim Gates, Phillip Farmer, Michael Rutherford, Geri Blake, et~al.,
\newblock ``Chest imaging representing a covid-19 positive rural us population,''
\newblock {\em Scientific data}, vol. 7, no. 1, pp. 414, 2020.

\bibitem{liu2020chestx}
Jingyu Liu, Jie Lian, and Yizhou Yu,
\newblock ``Chestx-det10: chest x-ray dataset on detection of thoracic abnormalities,''
\newblock {\em arXiv preprint arXiv:2006.10550}, 2020.

\bibitem{siim-acr-pneumothorax-segmentation}
Carol~Wu Anna~Zawacki, Julia~Elliott George~Shih, ParasLakhani Mikhail~Fomitchev, Mohannad~Hussain, and Shunxing~Bao Phil~Culliton,
\newblock ``Siim-acr pneumothorax segmentation,'' \url{https://kaggle.com/competitions/siim-acr-pneumothorax-segmentation}, 2019.

\bibitem{saporta2022benchmarking}
Adriel Saporta, Xiaotong Gui, Ashwin Agrawal, Anuj Pareek, Steven~QH Truong, Chanh~DT Nguyen, Van-Doan Ngo, Jayne Seekins, Francis~G Blankenberg, Andrew~Y Ng, et~al.,
\newblock ``Benchmarking saliency methods for chest x-ray interpretation,''
\newblock {\em Nature Machine Intelligence}, vol. 4, no. 10, pp. 867--878, 2022.

\bibitem{wang2017chestx}
Xiaosong Wang, Yifan Peng, Le~Lu, Zhiyong Lu, Mohammadhadi Bagheri, and Ronald~M Summers,
\newblock ``Chestx-ray8: Hospital-scale chest x-ray database and benchmarks on weakly-supervised classification and localization of common thorax diseases,''
\newblock in {\em Proceedings of the IEEE conference on computer vision and pattern recognition}, 2017, pp. 2097--2106.

\bibitem{nguyen2022vindr}
Ha~Q Nguyen, Khanh Lam, Linh~T Le, Hieu~H Pham, Dat~Q Tran, Dung~B Nguyen, Dung~D Le, Chi~M Pham, Hang~TT Tong, Diep~H Dinh, et~al.,
\newblock ``Vindr-cxr: An open dataset of chest x-rays with radiologist’s annotations,''
\newblock {\em Scientific Data}, vol. 9, no. 1, pp. 429, 2022.

\bibitem{zhang2023knowledge}
Xiaoman Zhang, Chaoyi Wu, Ya~Zhang, Weidi Xie, and Yanfeng Wang,
\newblock ``Knowledge-enhanced visual-language pre-training on chest radiology images,''
\newblock {\em Nature Communications}, vol. 14, no. 1, pp. 4542, 2023.

\bibitem{luo2024devide}
Haozhe Luo, Ziyu Zhou, Corentin Royer, Anjany Sekuboyina, and Bjoern Menze,
\newblock ``Devide: Faceted medical knowledge for improved medical vision-language pre-training,''
\newblock {\em arXiv preprint arXiv:2404.03618}, 2024.

\bibitem{wang2022medclip}
Zifeng Wang, Zhenbang Wu, Dinesh Agarwal, and Jimeng Sun,
\newblock ``Medclip: Contrastive learning from unpaired medical images and text,''
\newblock {\em arXiv preprint arXiv:2210.10163}, 2022.

\bibitem{luo2024dwarf}
Haozhe Luo, Aur{\'e}lie~Pahud de~Mortanges, Oana Inel, and Mauricio Reyes,
\newblock ``Dwarf: Disease-weighted network for attention map refinement,''
\newblock in {\em International Conference on Medical Image Computing and Computer-Assisted Intervention}. Springer, 2024, pp. 59--68.

\bibitem{pahud2024orchestrating}
Aur{\'e}lie Pahud~de Mortanges, Haozhe Luo, Shelley~Zixin Shu, Amith Kamath, Yannick Suter, Mohamed Shelan, Alexander P{\"o}llinger, and Mauricio Reyes,
\newblock ``Orchestrating explainable artificial intelligence for multimodal and longitudinal data in medical imaging,''
\newblock {\em NPJ digital medicine}, vol. 7, no. 1, pp. 195, 2024.

\bibitem{radford2021learning}
Alec Radford, Jong~Wook Kim, Chris Hallacy, Aditya Ramesh, Gabriel Goh, Sandhini Agarwal, Girish Sastry, Amanda Askell, Pamela Mishkin, Jack Clark, et~al.,
\newblock ``Learning transferable visual models from natural language supervision,''
\newblock in {\em International conference on machine learning}. PMLR, 2021, pp. 8748--8763.

\bibitem{zhang2023biomedclip}
Sheng Zhang, Yanbo Xu, Naoto Usuyama, Hanwen Xu, Jaspreet Bagga, Robert Tinn, Sam Preston, Rajesh Rao, Mu~Wei, Naveen Valluri, et~al.,
\newblock ``Biomedclip: a multimodal biomedical foundation model pretrained from fifteen million scientific image-text pairs,''
\newblock {\em arXiv preprint arXiv:2303.00915}, 2023.

\bibitem{wu2023medklip}
Chaoyi Wu, Xiaoman Zhang, Ya~Zhang, Yanfeng Wang, and Weidi Xie,
\newblock ``Medklip: Medical knowledge enhanced language-image pre-training for x-ray diagnosis,''
\newblock in {\em Proceedings of the IEEE/CVF International Conference on Computer Vision}, 2023, pp. 21372--21383.

\bibitem{phan2024decomposing}
Vu~Minh~Hieu Phan, Yutong Xie, Yuankai Qi, Lingqiao Liu, Liyang Liu, Bowen Zhang, Zhibin Liao, Qi~Wu, Minh-Son To, and Johan~W Verjans,
\newblock ``Decomposing disease descriptions for enhanced pathology detection: A multi-aspect vision-language pre-training framework,''
\newblock in {\em Proceedings of the IEEE/CVF Conference on Computer Vision and Pattern Recognition}, 2024, pp. 11492--11501.

\bibitem{lai2024carzero}
Haoran Lai, Qingsong Yao, Zihang Jiang, Rongsheng Wang, Zhiyang He, Xiaodong Tao, and S~Kevin Zhou,
\newblock ``Carzero: Cross-attention alignment for radiology zero-shot classification,''
\newblock in {\em Proceedings of the IEEE/CVF Conference on Computer Vision and Pattern Recognition}, 2024, pp. 11137--11146.

\end{thebibliography}

\end{document}